\def\year{2021}\relax
\documentclass[letterpaper]{article} 
\usepackage{aaai21}  
\usepackage{times}  
\usepackage{helvet} 
\usepackage{courier}  
\usepackage[hyphens]{url}  
\usepackage{graphicx} 
\usepackage{graphicx}  
\usepackage{natbib}  
\usepackage{caption} 
\usepackage{diagbox}
\usepackage{multirow}
\usepackage{amsfonts}
\usepackage{amsmath}
\usepackage{floatrow}
\usepackage[dvipsnames]{xcolor}
\definecolor{urlcolor}{HTML}{00F9DE}

\newfloatcommand{capbtabbox}{table}[][\FBwidth]
\usepackage[switch]{lineno}  
\graphicspath{{figure/}}

\title{CardioGAN: Attentive Generative Adversarial Network with Dual Discriminators for Synthesis of ECG from PPG}

\author{Pritam Sarkar, Ali Etemad \\}

\affiliations{Dept. of ECE \& Ingenuity Labs Research Institute \\
Queen's University, Kingston, Canada \\
\{pritam.sarkar, ali.etemad\}@queensu.ca    
}

\begin{document}
\maketitle

\setlength{\belowdisplayskip}{0pt} 
\setlength{\belowdisplayshortskip}{0pt}
\setlength{\abovedisplayskip}{0pt} 
\setlength{\abovedisplayshortskip}{0pt}
\setlength{\belowcaptionskip}{0pt}

\begin{abstract}
Electrocardiogram (ECG) is the electrical measurement of cardiac activity, whereas Photoplethysmogram (PPG) is the optical measurement of volumetric changes in blood circulation. 
While both signals are used for heart rate monitoring, from a medical perspective, ECG is more useful as it carries additional cardiac information. Despite many attempts toward incorporating ECG sensing in smartwatches or similar wearable devices for continuous and reliable cardiac monitoring, PPG sensors are the main feasible sensing solution available. 
In order to tackle this problem, we propose CardioGAN, an adversarial model which takes PPG as input and generates ECG as output. The proposed network utilizes an attention-based generator to learn local salient features, as well as dual discriminators to preserve the integrity of generated data in both time and frequency domains. Our experiments show that the ECG generated by CardioGAN provides more reliable heart rate measurements compared to the original input PPG, reducing the error from $9.74$ beats per minute (measured from the PPG) to $2.89$ (measured from the generated ECG).

\end{abstract}

\section{Introduction}
According to the World Health Organization (WHO) in 2017, Cardiovascular Deceases (CVDs) are reported as the leading causes of death worldwide \cite{who_cvds}. The report indicates that CVDs cause $31\%$ of global deaths, out of which at least three-quarters of deaths occur in the low or medium-income countries. One of the primary reasons behind this is the lack of primary healthcare support and the inaccessible on-demand health monitoring infrastructure. Electrocardiogram (ECG) is considered as one of the most important attributes for continuous health monitoring required for identifying those who are at serious risk of future cardiovascular events or death. Vast amount of research is being conducted with the goal of developing wearable devices capable of continuous ECG monitoring and feasible for daily life use, largely to no avail. Currently, very few wearable devices provide wrist-based ECG monitoring, and those who do, require the user to stand still and touch the watch with both hands in order to \textit{close the circuit} in order to record an ECG segment of limited duration (usually 30 seconds), making these solutions non-continuous and sporadic.

Photoplethysmogram (PPG), an optical method for measuring blood volume changes at the surface of the skin, is considered as a close alternative to ECG, which contains valuable cardiovascular information \cite{gil2010photoplethysmography,schafer2013accurate}. For instance, studies have shown that a number of features extracted from PPG (e.g., pulse rate variability) are highly correlated with corresponding metrics extracted from ECG (e.g., heart rate variability) \cite{gil2010photoplethysmography}, further illustrating the mutual information between these two modalities. Yet, through recent advancements in smartwatches, smartphones, and other similar wearable and mobile devices, PPG has become the industry standard as a simple, wearable-friendly, and low-cost solution for continuous heart rate (HR) monitoring for everyday use. Nonetheless, PPG suffers from inaccurate HR estimation and several other limitations in comparison to conventional ECG monitoring devices \cite{bent2020investigating} due to factors like skin tone, diverse skin types, motion artefacts, and signal crossovers among others. Moreover, the ECG waveform carries important information about cardiac activity. For instance, the P-wave indicates the sinus rhythm, whereas a long PR interval is generally indicative of a first-degree heart blockage \cite{ashley2004conquering}. As a result, ECG is consistently being used by cardiologists for assessing the condition and performance of the heart.

Based on the above, there is a clear discrepancy between the need for continuous wearable ECG monitoring and the available solutions in the market. To address this, we propose CardioGAN, a generative adversarial network (GAN) \cite{goodfellow2014generative}, which takes PPG as inputs and generates ECG. Our model is based on the CycleGAN architecture \cite{CycleGAN2017} which enables the system to be trained in an unpaired manner. Unlike CycleGAN, CardioGAN is designed with attention-based generators and equipped with multiple discriminators. We utilize attention mechanisms in the generators to better learn to focus on specific local regions such as the QRS complexes of ECG. To generate high fidelity ECG signals in terms of both time and frequency information, we utilize a dual discriminator strategy where one discriminator operates on signals in the time domain while the other uses frequency-domain spectrograms of the signals. We show that the generated ECG outputs are very similar to the corresponding real ECG signals. Finally, we perform HR estimation using our generated ECG as well as the input PPG signals. By comparing these values to the HR measured from the ground-truth ECG signals, we observe a clear advantage in our proposed method. While to demonstrate the efficacy of our solution we focus on single-lead ECG, we believe our approach can be used for multi-lead ECG through training the system on other desired leads. Our contributions in this paper are summarised below:
\begin{itemize}
    \item We propose a novel framework called CardioGAN for generating ECG signals from PPG inputs. We utilize attention-based generators and dual time and frequency domain discriminators along with a CycleGAN backbone to obtain realistic ECG signals.
    To the best of our knowledge, no other studies have attempted to generate ECG from PPG (or in fact any cross-modality signal-to-signal translation in the biosignal domain) using GANs or other deep learning techniques. 
    \item We perform a multi-corpus subject-independent study, which proves the generalizability of our model to data from unseen subjects and acquired in different conditions.
    \item The generated ECG obtained from the CardioGAN provides more accurate HR estimation compared to HR values calculated from the original PPG, demonstrating some of the benefits of our model in the healthcare domain. We make the final trained model publicly available\footnote{\url{https://code.engineering.queensu.ca/17ps21/ppg2ecg-cardiogan}}. 
\end{itemize}

The rest of this paper is organized as follows. Section \ref{Related Work} briefly mentions the prior studies on ECG signal generation. Next, our proposed method is discussed in Section \ref{Method}. Section \ref{experiments} discusses the details of our experiments, including datasets and training procedures. Finally, the results and analyses are presented in Section \ref{performance}, followed by a summary of our work in Section \ref{conclusion}.

\section{Related Work} \label{Related Work}

\subsection{Generating synthetic ECG Signal}
The idea of synthesizing ECG has been explored in the past, utilizing both model-driven (e.g. signal processing or mathematical modelling) and data-driven (machine learning and deep learning) techniques. As examples of earlier works, \cite{mcsharry2003dynamical,sayadi2010synthetic} proposed solutions based on differential equations and Gaussian models for generating ECG segments.

Despite deep learning being employed to process ECG for a wide variety of different applications, for instance biometrics \cite{heartid}, arrhythmia detection \cite{andrewng_nature}, emotion recognition \cite{sarkar2020,sarkar2019self}, cognitive load analysis \cite{sarkar2019,ross2019toward}, and others, very few studies have tackled \textit{synthesis} of ECG signals with deep neural networks \cite{zhu2019electrocardiogram,golany2019pgans,golany2020improving}.
Synthesizing ECG with GANs was first studied in \cite{zhu2019electrocardiogram}, where a bidirectional LSTM-CNN architecture was proposed to generate ECG from Gaussian noise. The study performed by \cite{golany2019pgans}, proposed PGAN or Personalized GAN to generate patient-specific synthetic ECG signals from input noise. A special loss function was proposed to mimic the morphology of ECG waveforms, which was a combination of cross-entropy loss and mean squared error between real and fake ECG waveforms. 

A few other studies have targeted this area, for example, EmotionalGAN was proposed in \cite{chen2019emotionalgan}, where synthetic ECG was used to augment the available ECG data in order to improve emotion classification accuracy. The proposed GAN generated the new ECG based on input noise. Lastly, in a similar study performed by \cite{golany2020improving}, ECG was generated from input noise to augment the available ECG training set, improving the performance for arrhythmia detection.

\subsection{ECG Synthesis from PPG}
With respect to the very specific problem of PPG-to-ECG translation, to the best of our knowledge, only \cite{zhu2019learning} has been published. This work did not use deep learning, instead used discrete cosine transformation (DCT) technique to map each PPG cycle to its corresponding ECG cycle. First, onsets of the PPG signals were aligned to the R-peaks of the ECG signals, followed by a de-trending operation in order to reduce noise. Next, each cycle of ECG and PPG was segmented, followed by temporal scaling using linear interpolation in order to maintain a fixed segment length. Finally, a linear regression model was trained to learn the relation between DCT coefficients of PPG segments and corresponding ECG segments. In spite of several contributions, this study suffers from few limitations. First, the model failed to produce reliable ECG in a subject-independent manner, which limits its application to only previously seen subject's data. Second, often the relation between PPG segments and ECG segments are not linear, therefore in several cases, this model failed to capture the non-linear relationships between these $2$ domains. Lastly, no experiments have been performed to indicate any performance enhancement gained from using the generated ECG as opposed to the available PPG (for example a comparison of measured HR).

\section{Method} \label{Method}
\subsection{Objective and Proposed Architecture}
In order to not be constrained by paired training where both types of data are needed from the same instance in order to train the system, we are interested in an unpaired GAN, i.e. CycleGAN-based architectures. We propose CardioGAN whose main objective is to learn to estimate the mapping between PPG ($P$) and ECG ($E$) domains. 
In order to force the generator to focus on regions of the data with significant importance, we incorporate an attention mechanism into the generator. We implement generator $G_{E}: P \rightarrow E$ to learn forward mapping, and $G_{P}: E \rightarrow P$ to learn the inverse mapping. We denote generated ECG and generated PPG from CardioGAN as $E'$ and $P'$ respectively, where $E' = G_{E}(P)$ and $P' = G_{P}(E)$. According to \cite{penttila2001time} and a large number of other studies, cardiac activity is manifested in both time and frequency domains. Therefore, in order to preserve the integrity of the generated ECG in both domains, we propose the use of a dual discriminator strategy, where $D^t$ is employed to classify time domain and $D^f$ is used to classify the frequency domain response of real and generated data. 

\begin{figure}[t!]
    \centering
    \includegraphics[width=\columnwidth]{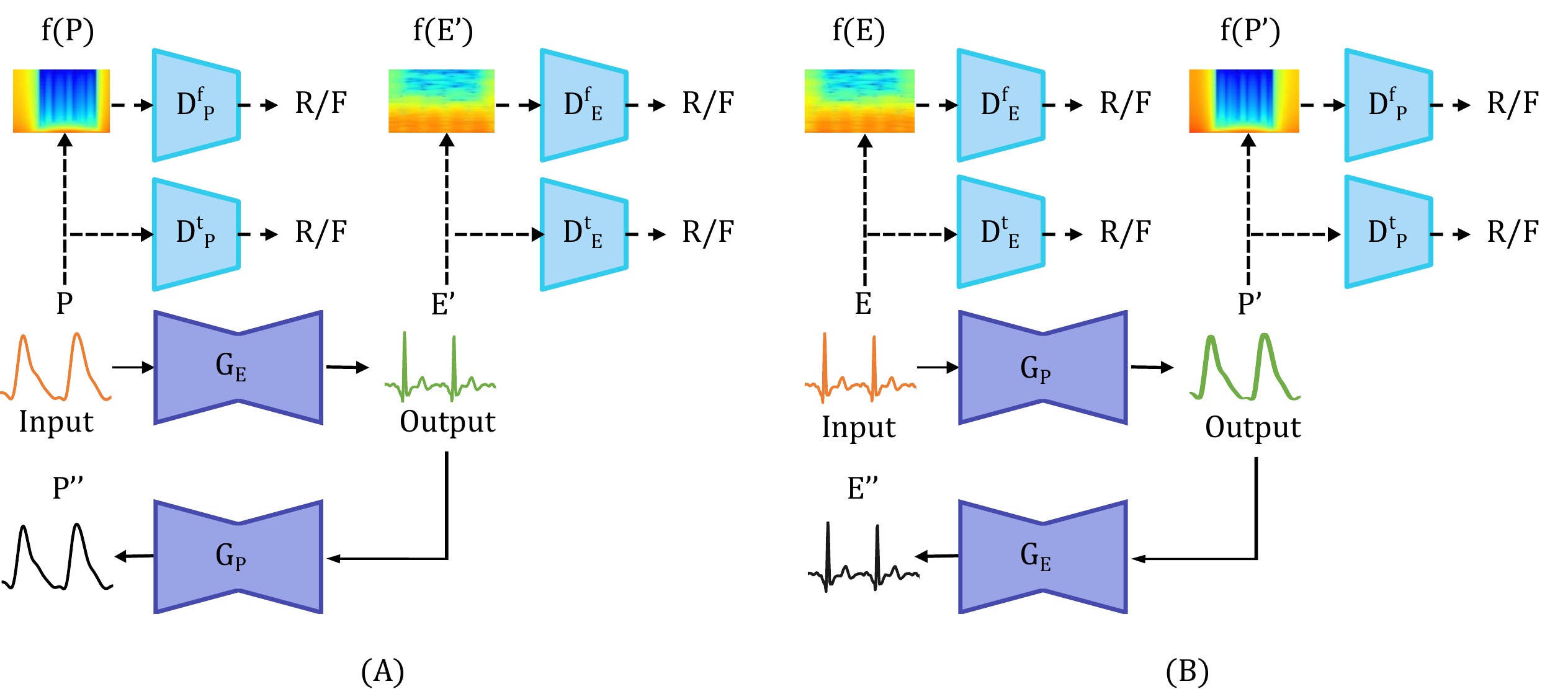}
    \caption{The architecture of the proposed CardioGAN is presented. The original ECG ($E$) and PPG ($P$) signals are shown in the color `orange'; the generated outputs ($E'$ and $P'$) are represented with the color `green'; and the reconstructed or cyclic outputs ($E''$ and $P''$) are marked with the color `black' for better visibility. Moreover, connections to the generators are marked with solid lines, whereas, connections to the discriminators are marked with dashed lines.}
    \label{fig:cardiogan}
\end{figure}

Figure \ref{fig:cardiogan} shows our proposed architecture, where $G_E$ takes $P$ as an input and generates $E'$ as the output. Similarly, $E$ is given as an input to $G_P$ where $P'$ is generated as the output. We employ $D^t_{E}$ and $D^t_{P}$ to discriminate $E$ versus $E'$, and $P$ versus $P'$, respectively. Similarly, $D^f_{E}$ and $D^f_{P}$ are developed to discriminate $f(E)$ versus $f(E')$, as well as $f(P)$ versus $f(P')$, respectively, where $f$ denotes the spectrogram of the input signal. Finally, $E'$ and $P'$ are given as inputs to $G_P$ and $G_E$ respectively, in order to complete the cyclic training process. 

In the following subsections, we expand on the dual discriminator, the notion of integrating an attention mechanism into the generator, and the loss functions used to train the overall architecture. The details and architectures of each of the networks used in our proposed solution are provided in Section \ref{architecture}.

\subsection{Dual Discriminators}

As mentioned above, to preserve both time and frequency information in the generated ECG, we use a dual discriminator approach. Dual discriminators have been used earlier in \cite{nguyen2017dual}, showing improvements in dealing with mode collapse problems. To leverage the concept of dual discriminators, we perform Short-Time Fourier Transformation (STFT) on the ECG/PPG time series data. Let's denote $x[n]$ as a time-series, then $STFT(x[n])$ can be denoted as $X(m, \omega) = \sum^\infty_{n=-\infty} x[n]w[n-m]e^{-j\omega n}$, where $m$ is the step size and $w[n]$ denotes \textit{Hann} window function.
Finally, the spectrogram is obtained by $f(x[n]) = log(|X(m, \omega)|+ \theta)$, where we use $\theta = 1e^{-10}$ to avoid infinite condition. As shown in Figure \ref{fig:cardiogan} the time-domain and frequency-domain discriminators operate in parallel, and as we will discuss in Section \ref{loss_sec}, to aggregate the outcomes of these two networks, the loss terms of both of these networks are incorporated into the adversarial loss.

\subsection{Attention-based Generators}
We adopt Attention U-Net as our generator architecture, which has been recently proposed and used for image classification \cite{oktay2018attention,jetley2018learn}. We chose attention-based generators to learn to better focus on salient features passing through the skip connections. Let's assume $x^l$ are features obtained from the skip connection originating from layer $l$, and $g$ is the gating vector that determines the region of focus. First, $x^l$ and $g$ are mapped to an intermediate-dimensional space $\mathbb{R}^{F_{int}}$ where $F_{int}$ corresponds to the dimensions of the intermediate-dimensional space. 
Our objective is to determine the scalar attention values ($\alpha_i^l$) for each temporal unit $x^l_i \in \mathbb{R}^{F_{l}}$, utilizing gating vector $g_i \in \mathbb{R}^{F_g}$, where $F_l$ and $F_g$ are the number of feature maps in $x^l$ and $g$ respectively. Linear transformations are performed on $x^l$ and $g$ as $\theta_x = W_x x^l_i + b_x$ and $\theta_g = W_g g_i + b_g$ respectively, where  $W_x \in  \mathbb{R}^{F_{l} \times F_{int}}$, $W_g \in  \mathbb{R}^{F_{g} \times F_{int}}$, and $b_x$, $b_g$ refer to the bias terms. Next, non-linear activation function \textit{ReLu} (denoted by $\sigma_1$) is applied to obtain the sum feature activation $f = \sigma_1( \theta_x + \theta_g)$, where  $\sigma_1(y)$ is formulated as $\max(0, y)$. Next we perform a linear mapping of $f$ onto the $\mathbb{R}^{F_{int}}$ dimensional space by performing channel-wise $1 \times 1$ convolutions, followed by passing through a sigmoid activation function $(\sigma_2)$ in order to obtain the attention weights in the range of $[0, 1]$. The attention map corresponding to $x^l$ is obtained by $\alpha_{i}^l = \sigma_2(\psi * f)$ where $\sigma_2(y)$ can be formulated as $\frac{1}{1+exp^{-y}}$,  $\psi \in \mathbb{R}^{F_{int}\time 1}$ and $*$ denotes convolution. Next, we perform element-wise multiplication between $x^l_i$ and  $\alpha^l_i$ to obtain the final output from the attention layer.

\subsection{Loss} \label{loss_sec}

Our final objective function is a combination of an adversarial loss and a cyclic consistency loss as presented below. 

\subsubsection{Adversarial Loss}
We apply adversarial loss in both forward and inverse mappings. Let's denote individual PPG segments as $p$ and the corresponding ground-truth ECG segments as $e$. For the mapping function $G_{E}: P \rightarrow E$, and discriminators $D^t_{E}$ and $D^f_{E}$, the adversarial losses are defined as: 
\begin{equation}
\begin{split}
\mathcal{L}_{adv}(G_{E},D^t_{E}) & = \mathbb{E}_{e \sim E} [\log{(D^t_{E}(e))}] \\
& + \mathbb{E}_{p \sim P} [\log{(1- D^t_{E}(G_{E}(p)))}]  
\end{split}
\end{equation}
\begin{equation} 
\begin{split}
\mathcal{L}_{adv}(G_{E},D^f_{E}) & = \mathbb{E}_{e \sim E} [\log{(D^f_{E}(f(e)))}] \\ 
& + \mathbb{E}_{p \sim P} [\log{(1 - D^f_{E}(f(G_{E}(p))))}] 
\end{split}
\end{equation}

Similarly, for the inverse mapping function $G_{P}: E \rightarrow P$, and discriminators $D^t_{P}$ and $D^f_{P}$, the adversarial losses are defined as: 
\begin{equation}
\begin{split}
\mathcal{L}_{adv}(G_{P},D^t_{P}) & = \mathbb{E}_{p \sim P} [\log{(D^t_{P}(p))}] \\
& + \mathbb{E}_{e \sim E} [\log{(1- D^t_{P}(G_{P}(e)))}]  
\end{split}
\end{equation}
\begin{equation} 
\begin{split}
\mathcal{L}_{adv}(G_{P},D^f_{P}) & = \mathbb{E}_{p \sim P} [\log{(D^f_{P}(f(p)))}] \\
& + \mathbb{E}_{e \sim E} [\log{(1 - D^f_{P}(f(G_{P}(e))))}] 
\end{split}
\end{equation}

Finally, the adversarial objective function for the mapping $G_{E}: P \rightarrow E$ is obtained as $\min_{G_E} \max_{D^t_E}\mathcal{L}_{adv}(G_{E},D^t_{E})$ and $\min_{G_E} \max_{D^f_E}\mathcal{L}_{adv}(G_{E},D^f_{E})$. Similarly, for the mapping $G_{P}: E \rightarrow P$, can be calculated as $\min_{G_P} \max_{D^t_P}\mathcal{L}_{adv}(G_{P},D^t_{P})$ and $\min_{G_P} \max_{D^f_P}\mathcal{L}_{adv}(G_{P},D^f_{P})$.

\subsubsection{Cyclic Consistency Loss}
The other component of our objective function is the cyclic consistency loss or reconstruction loss as proposed by \cite{CycleGAN2017}. In order to ensure that forward mappings and inverse mappings are consistent, i.e., $p \rightarrow G_E(p) \rightarrow G_P(G_E(p)) \approx p$, as well as $e \rightarrow G_P(e) \rightarrow G_E(G_P(e)) \approx e$, we minimize the cycle consistency loss calculated as:
\begin{equation} 
\begin{split}
\mathcal{L}_{cyclic}(G_E, G_P) & = \mathbb{E}_{e \sim E} [||G_E(G_P(e))-e||_1] \\
& + \mathbb{E}_{p \sim P} [||G_P(G_E(p))-p||_1]
\end{split}
\end{equation}
\subsubsection{Final Loss}
The final objective function of CardioGAN is computed as:
\begin{equation} 
\begin{split} 
\mathcal{L}_{CardioGAN} & = \alpha \mathcal{L}_{adv}(G_{E},D^t_{E}) + \alpha \mathcal{L}_{adv}(G_{P},D^t_{P}) \\
& + \beta \mathcal{L}_{adv}(G_{E},D^f_{E}) + \beta \mathcal{L}_{adv}(G_{P},D^f_{P})  \\ &+ \lambda \mathcal{L}_{cyclic}(G_E, G_P) ,
\end{split}
\end{equation}
where $\alpha$ and $\beta$ are adversarial loss coefficients corresponding to $D^t$ and $D^f$ respectively, and $\lambda$ is the cyclic consistency loss coefficient.

\section{Experiments} \label{experiments}
In this section, we first introduce the datasets used in this study, followed by the description of the data preparation steps. Next, we present our implementation and architecture details.

\subsection{Datasets}
We use $4$ very popular ECG-PPG datasets, namely BIDMC \cite{pimentel2016toward}, CAPNO \cite{karlen2013multiparameter}, DALIA \cite{reiss2019deep}, and WESAD \cite{schmidt2018introducing}. We combine these $4$ datasets in order to enable a multi-corpus approach leveraging large and diverse distributions of data for different factors such as activity (e.g. working, driving, walking, resting), age (e.g. $29$ children, $96$ adults), and others. The aggregate dataset contains a total of $125$ participants with a balanced male-female ratio.

\subsubsection{BIDMC} \cite{pimentel2016toward} was obtained from $53$ adult ICU patients ($32$ females, $21$ males, mean age of $64.81$) where each recording was $8$ minutes long. PPG and ECG were both sampled at a frequency of $125$ Hz. It should be noted this dataset consists of three leads of ECG (II, V, AVR). However, we only use lead II in this study.

\subsubsection{CAPNO} \cite{karlen2013multiparameter} consists of data from $42$ participants, out of which $29$ were children (median age of $8.7$) and $13$ were adults (median age of $52.4$). The recordings were collected while the participants were under medical observation. Single-lead ECG and PPG recordings were sampled at a frequency of $300$ Hz and were $8$ minutes in length. 

\subsubsection{DALIA} \cite{reiss2019deep} was recorded from $15$ participants ($8$ females, $7$ males, mean age of $30.60$), where each recording was approximately $2$ hours long. ECG and PPG signals were recorded while participants went through different daily life activities, for instance sitting, walking, driving, cycling, working and so on. Single-lead ECG signals were recorded at a sampling frequency of $700$ Hz while the PPG signals were recorded at a sampling rate of $64$ Hz. 

\subsubsection{WESAD} \cite{schmidt2018introducing} was created using data from $15$ participants ($12$ male, $3$ female, mean age of $27.5$), while performing activities such as solving arithmetic tasks, watching video clips, and others. Each recording was over $1$ hour in duration. Single-lead ECG was recorded at a sampling rate of $700$ Hz while PPG was recorded at a sampling rate of $64$ Hz.

\subsection{Data Preparation} 
Since the above-mentioned datasets have been collected at different sampling frequencies, as a first step we re-sampled (using interpolation) both the ECG and PPG signals with a sampling rate of $128$ Hz. 
As the raw physiological signals contain a varying amounts and types of noise (e.g. power line interference, baseline wandering, motion artefacts), we perform very common filtering techniques on both the ECG and PPG signals. We apply a band-pass FIR filter with a pass-band frequency of $3$ Hz and stop-band frequency of $45$ Hz on the ECG signals. Similarly, a band-pass Butterworth filter with a pass-band frequency of $1$ Hz and a stop-band frequency of $8$ Hz is applied on the PPG signals. Next, person-specific z-score normalization is performed on both ECG and PPG. Then, the normalized ECG and PPG signals are segmented into $4$-second windows  ($128$ Hz $\times 4$ seconds $= 512$ samples), with a $10\%$ overlap to avoid missing any peaks. 
Finally, we perform min-max $[-1, 1]$ normalization on both ECG and PPG segments to ensure all the input data are in a specific range.

\subsection{Architecture} \label{architecture}

\subsubsection{Generator}
As mentioned earlier an Attention U-Net architecture is used as our generator, where self-gated soft-attention units are used to filter the features passing through the skip connections. $G_{E}$ and $G_{P}$ take $1\times512$ data points as input. The encoder consists of $6$ blocks, where the number of filters is gradually increased ($64, 128, 256, 512, 512, 512$) with a fixed kernel size of $1\times16$ and a stride of $2$. We apply layer normalization and leaky-ReLu activation after each convolution layers except the first layer, where no normalization is used. A similar architecture is used in the decoder, except de-convolutional layers with ReLu activation functions are used and the number of filters is gradually decreased in the same manner. The final output is then obtained from a de-convolutional layer with a single-channel output followed by tanh activation.

\subsubsection{Discriminator} 
Dual discriminators are used to classify real and fake data in time and frequency domains. $D^t_{E}$ and $D^t_{P}$ take time-series signals of size $1\times512$ as inputs, whereas, spectrograms of size $128\times128$ are given as inputs to $D^f_{E}$ and $D^f_{P}$. Both $D^t$ and $D^f$ use $4$ convolution layers, where the number of filters are gradually increased ($64, 128, 256, 512$) with a fixed kernel of $1\times16$ for $D^t$ and $7\times7$ for $D^f$. Both networks use a stride of 2. Each convolution layer is followed by layer normalization and leaky ReLu activation, except the first layer where no normalization is used. Finally, the output is obtained from a single-channel convolutional layer.

\subsection{Training} \label{training}
Our proposed CardioGAN network is trained from scratch on an Nvidia Titan RTX GPU, using TensorFlow 2.2. We divide the aggregated dataset into a training set and test set. We randomly select $80\%$ of the users from each dataset (a total of $101$ participants, equivalent to $58$K segments) for training, and the remaining $20\%$ of users from each dataset (a total of $24$ participants, equivalent to $15$K segments) for testing. The training time was approximately $50$ hours. To enable CardioGAN to be trained in an \textit{unpaired} fashion, we shuffle the ECG and PPG segments from each dataset separately eliminating the couplings between ECG and PPG followed by a shuffling of the order of datasets themselves for ECG and PPG separately. We use a batch size of $128$, unlike the original CycleGAN where a batch size of $1$ is used. We notice performance gain with a larger batch size. Adam optimizer is used to train both the generators and discriminators. We train our model for $15$ epochs, where the learning rate ($1e^{-4}$) is kept constant for the initial $10$ epochs and then linearly decayed to $0$. The values of $\alpha$, $\beta$, and $\lambda$ are empirically set to $3$, $1$ and $30$ respectively. 

\section{Performance} \label{performance}
CardioGAN produces two main signal outputs, generated ECG ($E'$) and generated PPG ($P'$). As our goal is to generate the more important and elusive ECG, we utilize $E'$ and ignore $P'$ in the following experiments. In this section, we present the quantitative and qualitative results of our proposed CardioGAN network. Next, we perform an ablation study in order to understand the effects of the different components of the model. Further, we perform several analyses, followed by a discussion of potential applications using our proposed solution.

\subsection{Quantitative Results}
Heart rate is measured as number of beats per minutes (BPM) by dividing the length of ECG or PPG segments in seconds by the average of the peak intervals multiplied by 60 (seconds). Let's define the mean absolute error (MAE) metric for the heart rate (in BPM) obtained from a given ECG or PPG signal ($HR^{Q}$) with respect to a ground-truth HR ($HR^{GT}$) as
$ MAE_{HR} (Q) = {\frac{1}{N}} \sum_{i=1}^{N} |HR^{GT}_i - HR^{Q}_i|$, where $N$ is the number of segments for which the HR measurements have been obtained. In order to investigate the merits of CardioGAN, we measure $ MAE_{HR} (E')$, where $E'$ is the ECG generated by CardioGAN. We compare these MAE values to $ MAE_{HR} (P)$ (where $P$ denotes the available input PPG) as reported by other studies on the 4 datasets. The results are presented in Table \ref{tab:bpm_comparison} where we observe that for $3$ of the $4$ datasets, the HR measured from the ECG generated by CardioGAN is more accurate than the HR measured from the input PPG signals. For CAPNO dataset in which our ECG shows higher error compared to other works based on PPG, the difference is quite marginal, especially in comparison to the performance gains achieved across the other datasets. 

Different studies in this area have used different window sizes for HR measurement which we report in Table \ref{tab:bpm_comparison}. To evaluate the impact of our solution based on different window sizes, we measure $MAE_{HR} (E')$ over different $4, 8, 16, 32$, and $64$ second windows and present the results in comparison to $MAE_{HR} (P)$ across all the subjects available in the $4$ datasets in Table \ref{tab:bpm}. In these experiments, we utilize two popular algorithms for detecting peaks from ECG \cite{hamilton2002open} and PPG \cite{elgendi2013systolic} signals. We observe a clear advantage in measuring HR from $E'$ as opposed to $P$. We notice a very consistent performance gain across different window sizes, which further demonstrates the stability of the results produce by CardioGAN.

\begin{table}[t!]
\fontsize{9pt}{10pt}\selectfont
\setlength\tabcolsep{4pt}
\centering
\begin{tabular}{| l |c |c |c| }
\hline
\textbf{Dataset} & \textbf{Method} & \textbf{Window (sec.)} & $MAE_{HR}$     \\ \hline \hline
\multirow{6}{*}{\textbf{BIDMC}}    
      & \cite{nilsson2005respiration} &\multirow{6}{*}{$64$}& $4.6$ \\
      & \cite{shelley2006use} && $2.3$ \\
      & \cite{fleming2007comparison}  && $5.5$ \\
      & \cite{karlen2013multiparameter} && $5.7$     \\
      & \cite{pimentel2016toward} && $2.7$    \\ 
      & \textbf{CardioGAN}                 && $\textbf{0.7}$ \\ \hline
\multirow{6}{*}{\textbf{CAPNO}}
      & \cite{nilsson2005respiration} &\multirow{6}{*}{$64$}& $10.2$ \\
      & \cite{shelley2006use} && $2.2$ \\
      & \cite{fleming2007comparison}  && $1.4$ \\
      & \textbf{\cite{karlen2013multiparameter}} && $\textbf{1.2}$     \\
      & \cite{pimentel2016toward} && $1.9$    \\ 
      & CardioGAN                 && $2.0$ \\ \hline
\multirow{4}{*}{\textbf{Dalia}}    & \cite{schack2017computationally}  & \multirow{4}{*}{$8$} & $20.5$ \\
      & \cite{reiss2019deep} && $15.6$     \\
      & \cite{reiss2019deep} && $11.1$    \\ 
      & \textbf{CardioGAN}                 && $\textbf{8.3}$ \\ \hline
\multirow{4}{*}{\textbf{WESAD}}    & \cite{schack2017computationally}  & \multirow{4}{*}{$8$} & $19.9$ \\
      & \cite{reiss2019deep} && $11.5$     \\
      & \cite{reiss2019deep} && $9.5$    \\ 
      & \textbf{CardioGAN}                 && $\textbf{8.6}$ \\ \hline

\end{tabular}
\caption{We compare the $MAE_{HR}$ calculated from the generated ECG with $MAE_{HR}$ calculated from the real input PPG.}
\label{tab:bpm_comparison}
\end{table}

\begin{table}[t!]
\fontsize{9pt}{10pt}\selectfont
\centering
\begin{tabular}{|c |c |c| }
\hline
\textbf{Window (sec.)} & $MAE_{HR}(E')$ & $MAE_{HR}(P)$    \\ \hline \hline
         4  &  $4.86$ & $10.67$   \\
         8  &  $3.54$ & $10.23$   \\
         16 & $3.27$ &  $10.00$   \\
         32 & $3.08$ &  $9.77$    \\ 
         64 & $2.89$ &  $9.74$    \\ \hline
      
\end{tabular}
\caption{A comparison of $MAE_{HR}$ between generated ECG and real PPG is presented for different window sizes.}
\label{tab:bpm}
\end{table}

\begin{figure*}[htb!]
    \centering
    \includegraphics[width=\textwidth]{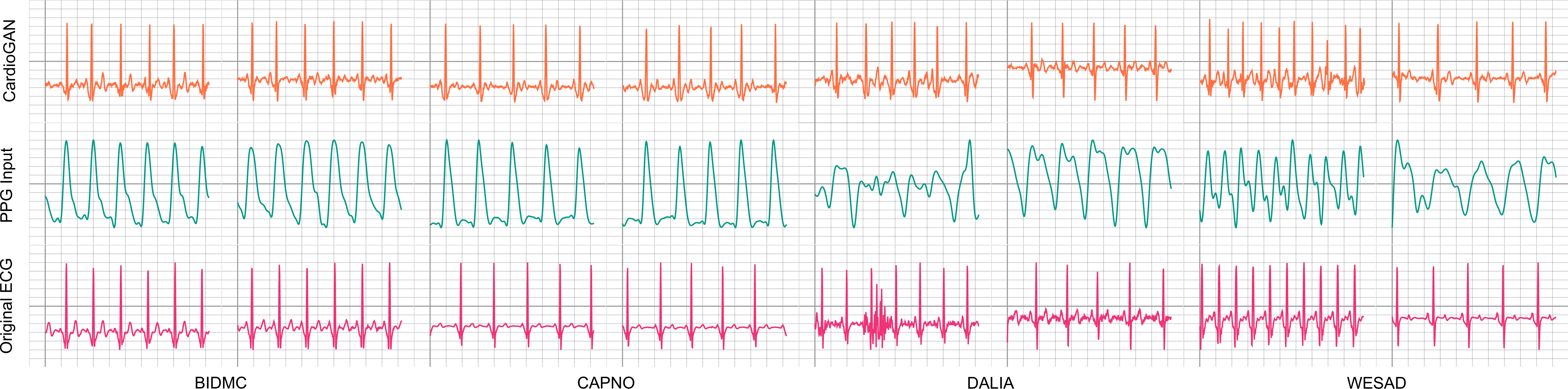}
    \caption{We present ECG samples generated by our proposed CardioGAN. We show $2$ different samples from each dataset to better demonstrate the qualitative performance of our method.}
    \label{fig:main_samples}
\end{figure*}

\subsection{Qualitative Results}
In Figure \ref{fig:main_samples} we present a number of samples of ECG signals generated by CardioGAN, clearly showing that our proposed network is able to learn to reconstruct the shape of the original ECG signals from corresponding PPG inputs. Careful observation shows that in some cases, the generated ECG signals exhibit a small time lag with respect to the original ECG signals. The root cause of this time delay is the Pulse Arrival Time (PAT), which is defined as the time taken by the PPG pulse to travel from the heart to a distal site (from where PPG is collected, for example, wrist, fingertip, ear, or others) \cite{elgendi2019use}. 
Nonetheless, this time-lag is consistent for all the beats across a single generated ECG signal as a simple offset, and therefore does not impact HR measurements or other cardiovascular-related metrics. This is further evidenced by the accurate HR measurements presented earlier in Tables \ref{tab:bpm_comparison} and \ref{tab:bpm}.

\subsection{Ablation Study}

The proposed CardioGAN consists of attention-based generators and dual discriminators, as discussed earlier. In order to investigate the usefulness of the attention mechanisms and dual discriminators, we perform an ablation study of $2$ variations of the network by removing each of these components individually. To evaluate these components, we perform the same $MAE_{HR}$ along with a number of other metrics to quantify the quality of ECG waveforms. We use metrics similar to those used in \cite{zhu2019electrocardiogram}, which are Root Mean Squared Error (RMSE), Percentage Root Mean Squared Difference (PRD), and Fr{\'e}chet Distance (FD). We briefly defined these metrics as follows: 

\textbf{RMSE:} In order to understand the stability between $E$ and $E'$, we calculate $RMSE = \sqrt{{\frac{1}{N}} \sum_{i=1}^{N} (E_i - E'_i)^2} $ where $E_i$ and $E'_i$ refer to the $i^{th}$ point of $E$ and $E'$ respectively.

\textbf{PRD:} To quantify the distortion between $E$ and $E'$, we calculate $PRD = \sqrt{ \frac{\sum_{i=1}^{N} (E_i - E'_i)^2}{\sum_{i=1}^{N} (E_i)^2} \times 100}$.

\textbf{FD:} Fr{\'e}chet distance \cite{alt1995computing} is calculated to measure the similarity between the $E$ and $E'$. While calculating the distance between two curves, this distance considers the location and order of the data points, hence, giving a more accurate measure of similarity between two time-series signals. Let's assume $E$, a discrete signal, can be expressed as a sequence of $\{e_1, e_2, e_3, \dots, e_N\}$, and similarly $E'$ can be expressed as $\{e'_1, e'_2, e'_3, \dots, e'_N\}$. We can create a $2$-D matrix $M$ of corresponding data points by preserving the order of sequence $E$ and $E'$, where $M \subseteq \{(e, e')|e \in E, e' \in E'\} $. The discrete Fr{\'e}chet distance of $E$ and $E'$ is calculated as $FD = \min_M \max_{(e,e')\in M}d(e,e')$,
where $d(e,e')$ denotes the Euclidean distance between corresponding samples of $e$ and $e'$.

The results of our ablation study are presented in Table \ref{tab:eval}. We present the performance of different variants of CardioGAN for all the subjects across all $4$ datasets. \textit{CardioGAN w/o DD} is the variant with only the time domain discriminator and no change in the generator architecture. \textit{CardioGAN w/o attn} is the variant where the generator does not contain an attention mechanism. The results presented in the table evidently show the benefit of using the proposed CardioGAN over it's ablation variants.

\begin{table}[t!]
    \fontsize{9pt}{10pt}\selectfont
    \setlength\tabcolsep{2pt}
    \centering
    \begin{tabular}{|l|c c c|c|}
    \hline
        \textbf{Method} & RMSE & PRD & FD & $MAE_{HR}$ \\ \hline \hline
         CardioGAN w/o DD   & $ 0.396 $ & $ 8.742 $ & $ 0.717 $ & $9.57$\\ 
         CardioGAN w/o Attn  & $ 0.386 $ & $ 8.393 $ & $ 0.773 $& $9.67$ \\ 
         \textbf{CardioGAN (proposed)} & $ \textbf{0.364} $ & $ \textbf{8.356} $ & $ \textbf{0.694} $ & $\textbf{4.77}$\\  \hline 
    \end{tabular}
    \caption{Performance comparison of CardioGAN and it's ablation variations across all the subjects of the $4$ datasets are presented.}
    \label{tab:eval}
\end{table}

\subsection{Analysis}
\subsubsection{Attention Map}
\begin{figure}
    \centering
    \includegraphics[width=\columnwidth]{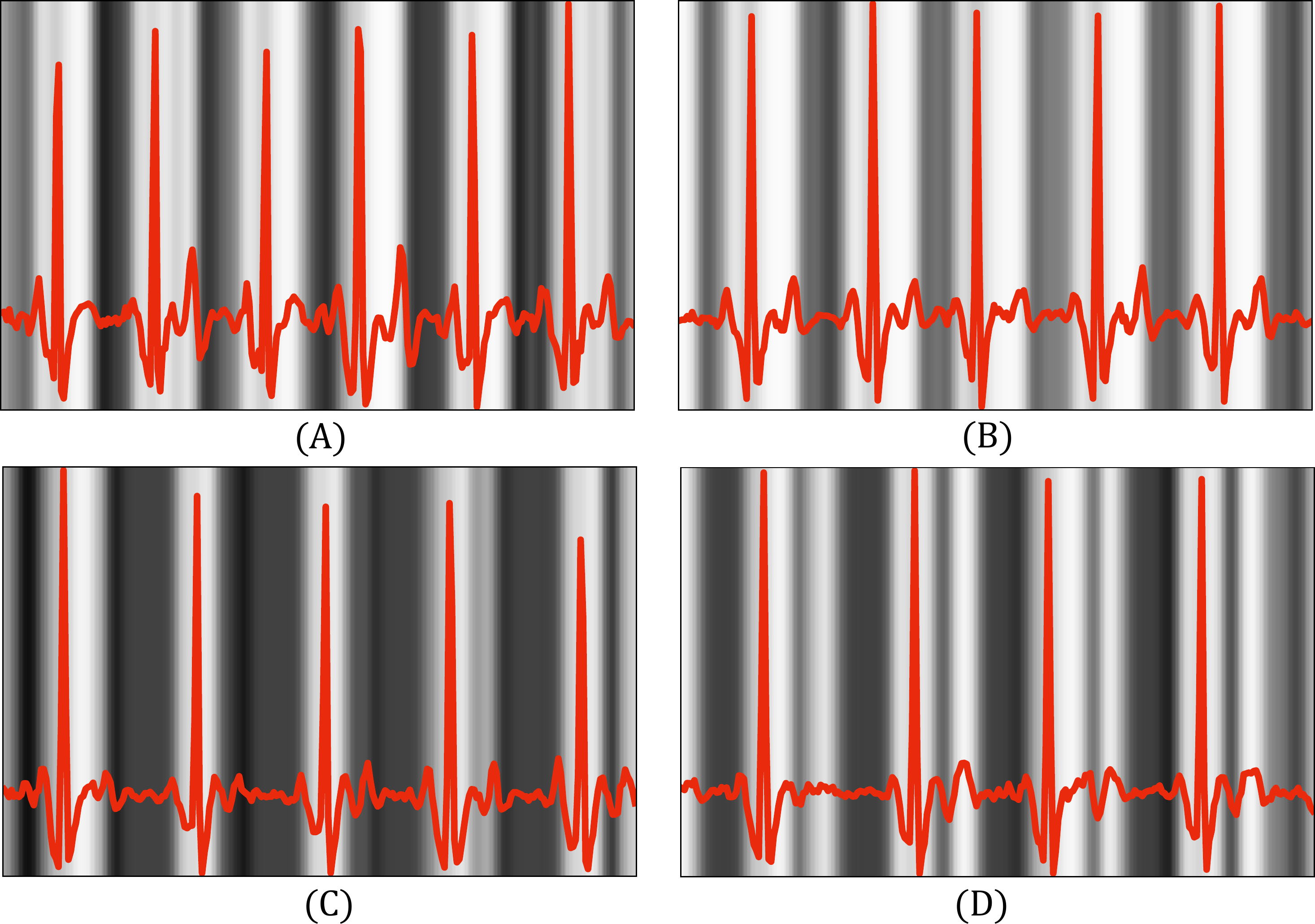}
    \caption{Visualization of attention maps are presented where the brighter parts indicate regions to which the generator pays more attention compared to the darker regions. We present $4$ samples of generated ECG segments corresponding to different subjects.}
    \label{fig:attn_map}
\end{figure}

In order to better understand what has been learned through the attention mechanism in the generators, we visualize the attention maps applied to the very last skip connection of the generator ($G_E$). We choose the attention applied to the last skip connection since this layer is the closest to the final output and there more interpretable. For better visualization, we superimpose the attention map on top of the output of the generator as shown in Figure \ref{fig:attn_map}. This shows that our model learns to generally focus on the PQRST complexes, which in turn helps the generator to learn the shapes of ECG waveform better as evident from qualitative and quantitative results presented earlier.

\subsubsection{Unpaired Training vs. Paired Training}
\begin{figure}[t!]
    \centering
    \includegraphics[width=\columnwidth]{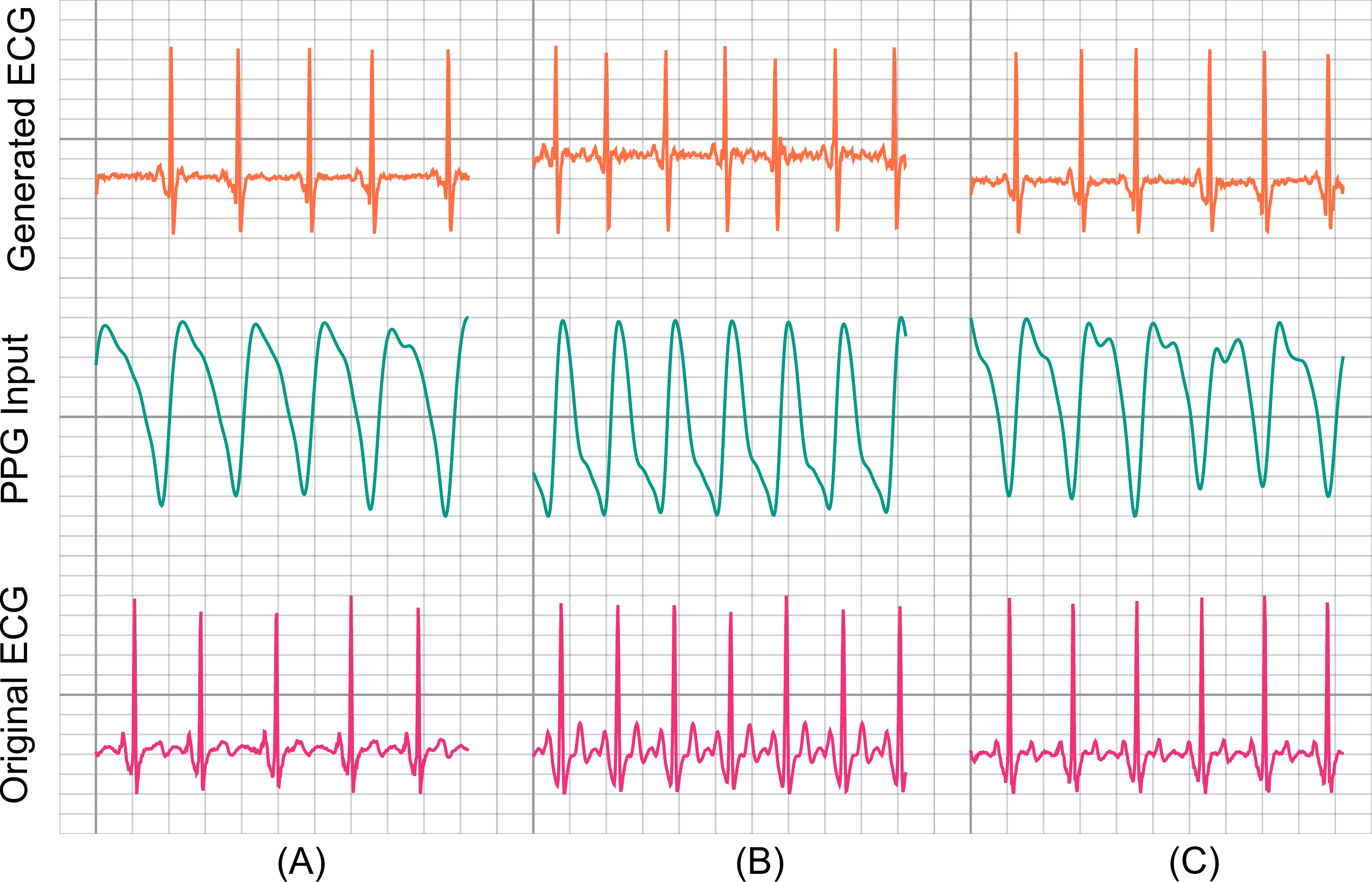}
    \caption{Samples obtained from paired training of CardioGAN are presented.}
    \label{fig:paired_samples}
\end{figure}

We further investigate the performance of CardioGAN while training with paired ECG-PPG inputs as opposed to our original approach which is based on unpaired training. To train CardioGAN in a paired manner, we follow the same training process mentioned in Section \ref{training}, except we keep the coupling between the ECG and PPG pairs intact in the input data. The results are presented in Table \ref{tab:eval_paired}, and a few samples of generated ECG are shown in Figure \ref{fig:paired_samples}. By comparing these results to those presented in Table \ref{tab:eval_paired}, we observe that unpaired training of CardioGAN shows superior performance compared to paired training. In particular, we notice that while CardioGAN-Paired does learns well to generate ECG beats from PPG inputs, it fails to learn the exact shape of the original ECG waveforms. This might be because an unpaired training scheme forces the network to learn stronger user-independent mappings between PPG and ECG, compared to user-dependant paired training. While it can be argued that utilizing paired data using other GAN architectures might perform well, it should be noted that the goal of this experiment is to evaluate the performance when paired training is performed without any fundamental changes to the architecture. We design CardioGAN with the aim of being able to leverage datasets that do not necessarily contain both ECG and PPG, hence, unpaired training, even though we resort to datasets that do contain both (ECG and PPG) so that ground-truth measurements can be used for evaluation purposes.

\begin{table}[t!]
    \fontsize{9pt}{10pt}\selectfont
    \setlength\tabcolsep{5pt}
    \centering
    \caption{The results obtained from CardioGAN-Paired are presented.}
    \begin{tabular}{| l |c c c| c|}
    \hline
         \multirow{1}{*}{\textbf{Method} } & RMSE & PRD & FD & $ MAE_{HR}$     \\ \hline \hline
         CardioGAN-Paired  & $0.437$  & $9.315$ & $0.748$ & $5.04$ \\ \hline 

    \end{tabular}
    \label{tab:eval_paired}
\end{table}

\begin{figure}[t!]
    \centering
    \includegraphics[width=\columnwidth]{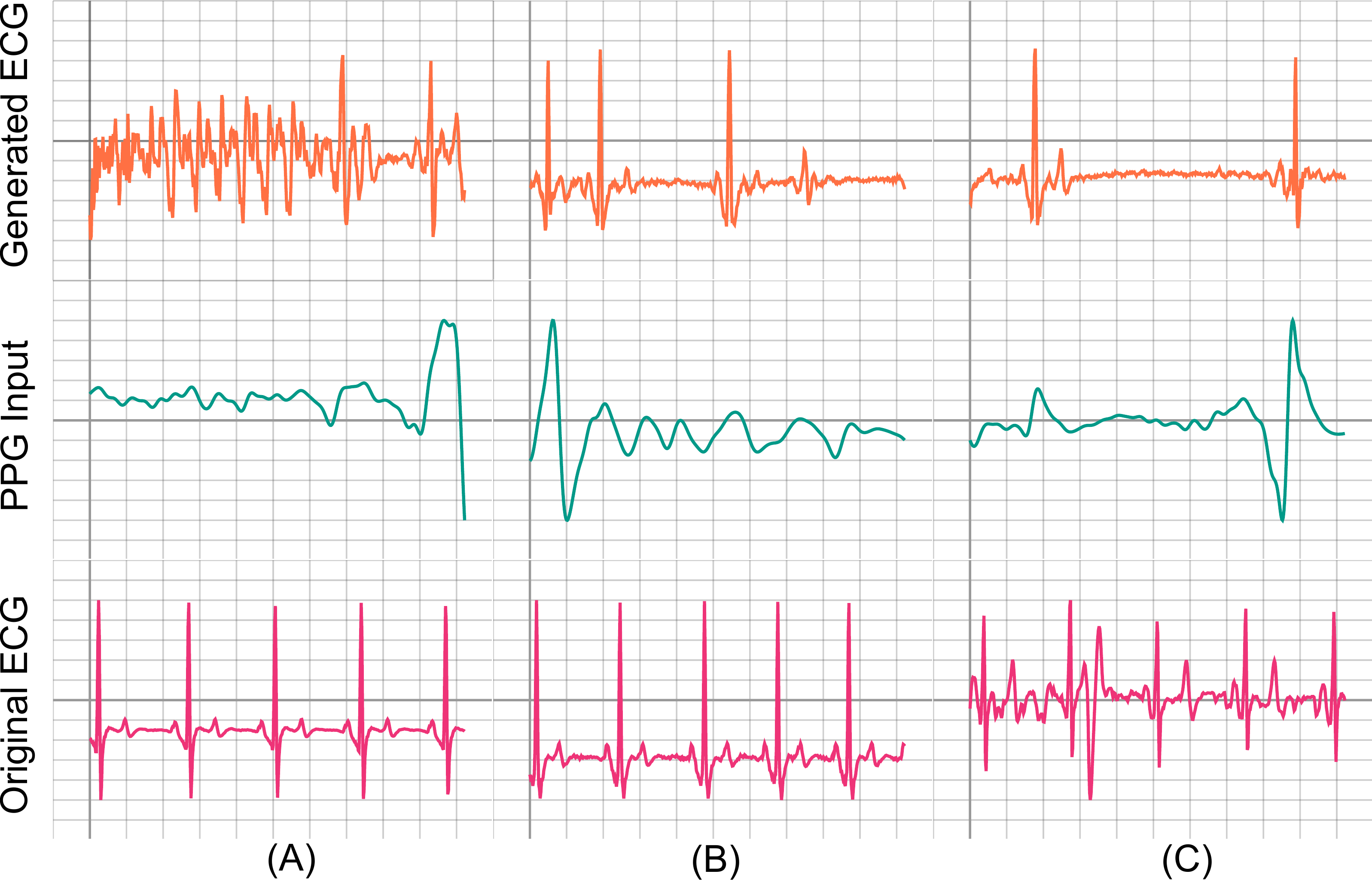}
    \caption{Few failed ECG examples generated by CardioGAN are presented.}
    \label{fig:bad_samples}
\end{figure}

\subsubsection{Failed Cases}
We notice there are instances where CardioGan fails to generate ECG samples that resemble the original ECG data very closely. Such cases arise only when the PPG input signals are of very poor quality. We show a few examples in Figure \ref{fig:bad_samples} where for highly noisy PPG inputs, the generated ECG samples also exhibit very low quality.

\subsection{Potential Applications and Demonstration}

Apart from the interest to the AI community, we believe our proposed solution has the potential to make a larger impact in the healthcare and wearable domains, notably for continuous health monitoring. Monitoring cardiac activity is an essential part of continuous health monitoring systems, which could enable early diagnosis of cardiovascular diseases, and in turn, early preventative measures that can lead to overcoming severe cardiac problems. Nonetheless, as discussed earlier, there are no suitable solutions for every-day continuous ECG monitoring. In this study we bridge this gap by utilizing PPG signals (which can be easily collected from almost every wearable devices available in the market) in our proposed CardioGAN to capture the cardiac information of users and generate accurate ECG signals. We perform a multi-corpus subject-independent study, where the subjects have gone through a wide range of activities including daily-life tasks, which assures us of the usability of our proposed solution in practical settings. Most importantly, our proposed solution can be integrated into an existing PPG-based wearable device to extract ECG data without any required additional hardware. To demonstrate this concept, we have implemented our model to perform in real-time and used a wrist-based wearable device to feed it with PPG data. The video\footnote{\url{https://youtu.be/z0Dr4k24t7U}} presented as supplementary material demonstrates CardioGAN producing realistic ECG from wearable PPG in real-time.


\section{Summary and Future Work} \label{conclusion}
In this paper, we propose CardioGAN, a solution for generating ECG signals from input PPG signals to aid with continuous and reliable cardiac monitoring. Our proposed method takes $4$-second PPG segments and generates corresponding ECG segments of equal length. Self-gated soft-attention is used in the generator to learn important regions, for example the QRS complexes of ECG waveforms. Moreover, a dual discriminator strategy is used to learn the mapping in both time and frequency domains. Further, we evaluate the merits of the generated ECG by calculating HR and comparing the results to HR obtained from the real PPG. The analysis shows a clear advantage of using CardioGAN as more accurate HR values are obtained as a result of using the model. 

For future work, the advantages of using the generated ECG data in other areas where the use of PPG is limited may be evaluated. These areas include identification of cardiovascular diseases, detection of abnormal heart rhythms, and others. Furthermore, generating multi-lead ECG can also be studied in order to extract more useful cardiac information often missing in single-channel ECG recordings. Finally, we hope our research can open a new path towards cross-modality signal-to-signal translation in the biosignal domain, allowing for less available physiological recording to be generated from more affordable and readily available signals.

\bibliography{references.bib}

\begin{thebibliography}{36}
\providecommand{\natexlab}[1]{#1}
\providecommand{\url}[1]{\texttt{#1}}
\providecommand{\urlprefix}{URL }
\expandafter\ifx\csname urlstyle\endcsname\relax
  \providecommand{\doi}[1]{doi:\discretionary{}{}{}#1}\else
  \providecommand{\doi}{doi:\discretionary{}{}{}\begingroup
  \urlstyle{rm}\Url}\fi

\bibitem[{Alt and Godau(1995)}]{alt1995computing}
Alt, H.; and Godau, M. 1995.
\newblock Computing the Fr{\'e}chet distance between two polygonal curves.
\newblock \emph{International Journal of Computational Geometry \&
  Applications} 5(01n02): 75--91.

\bibitem[{Ashley and Niebauer(2004)}]{ashley2004conquering}
Ashley, E.; and Niebauer, J. 2004.
\newblock \emph{Conquering the ECG}.
\newblock London: Remedica.

\bibitem[{Bent et~al.(2020)Bent, Goldstein, Kibbe, and
  Dunn}]{bent2020investigating}
Bent, B.; Goldstein, B.~A.; Kibbe, W.~A.; and Dunn, J.~P. 2020.
\newblock Investigating sources of inaccuracy in wearable optical heart rate
  sensors.
\newblock \emph{NPJ Digital Medicine} 3(1): 1--9.

\bibitem[{Chen et~al.(2019)Chen, Zhu, Hong, and Yang}]{chen2019emotionalgan}
Chen, G.; Zhu, Y.; Hong, Z.; and Yang, Z. 2019.
\newblock EmotionalGAN: Generating ECG to Enhance Emotion State Classification.
\newblock In \emph{Proceedings of the International Conference on Artificial
  Intelligence and Computer Science}, 309--313.

\bibitem[{Elgendi et~al.(2019)Elgendi, Fletcher, Liang, Howard, Lovell, Abbott,
  Lim, and Ward}]{elgendi2019use}
Elgendi, M.; Fletcher, R.; Liang, Y.; Howard, N.; Lovell, N.~H.; Abbott, D.;
  Lim, K.; and Ward, R. 2019.
\newblock The use of photoplethysmography for assessing hypertension.
\newblock \emph{NPJ Digital Medicine} 2(1): 1--11.

\bibitem[{Elgendi et~al.(2013)Elgendi, Norton, Brearley, Abbott, and
  Schuurmans}]{elgendi2013systolic}
Elgendi, M.; Norton, I.; Brearley, M.; Abbott, D.; and Schuurmans, D. 2013.
\newblock Systolic peak detection in acceleration photoplethysmograms measured
  from emergency responders in tropical conditions.
\newblock \emph{PLoS One} 8(10): e76585.

\bibitem[{Fleming et~al.(2007)}]{fleming2007comparison}
Fleming, S.~G.; et~al. 2007.
\newblock A comparison of signal processing techniques for the extraction of
  breathing rate from the photoplethysmogram.
\newblock \emph{International Journal of Biological and Medical Sciences} 2(4):
  232--236.

\bibitem[{Gil et~al.(2010)Gil, Orini, Bailon, Vergara, Mainardi, and
  Laguna}]{gil2010photoplethysmography}
Gil, E.; Orini, M.; Bailon, R.; Vergara, J.~M.; Mainardi, L.; and Laguna, P.
  2010.
\newblock Photoplethysmography pulse rate variability as a surrogate
  measurement of heart rate variability during non-stationary conditions.
\newblock \emph{Physiological Measurement} 31(9): 1271.

\bibitem[{Golany et~al.(2020)Golany, Lavee, Yarden, and
  Radinsky}]{golany2020improving}
Golany, T.; Lavee, G.; Yarden, S.~T.; and Radinsky, K. 2020.
\newblock Improving ECG Classification Using Generative Adversarial Networks.
\newblock In \emph{Proceedings of the AAAI Conference on Artificial
  Intelligence}, 13280--13285.

\bibitem[{Golany and Radinsky(2019)}]{golany2019pgans}
Golany, T.; and Radinsky, K. 2019.
\newblock PGANs: Personalized generative adversarial networks for ECG synthesis
  to improve patient-specific deep ECG classification.
\newblock In \emph{Proceedings of the AAAI Conference on Artificial
  Intelligence}, volume~33, 557--564.

\bibitem[{Goodfellow et~al.(2014)Goodfellow, Pouget-Abadie, Mirza, Xu,
  Warde-Farley, Ozair, Courville, and Bengio}]{goodfellow2014generative}
Goodfellow, I.; Pouget-Abadie, J.; Mirza, M.; Xu, B.; Warde-Farley, D.; Ozair,
  S.; Courville, A.; and Bengio, Y. 2014.
\newblock Generative adversarial nets.
\newblock In \emph{Advances in Neural Information Processing Systems},
  2672--2680.

\bibitem[{Hamilton(2002)}]{hamilton2002open}
Hamilton, P. 2002.
\newblock Open source ECG analysis.
\newblock In \emph{Computers in Cardiology}, 101--104. IEEE.

\bibitem[{Hannun et~al.(2019)Hannun, Rajpurkar, Haghpanahi, Tison, Bourn,
  Turakhia, and Ng}]{andrewng_nature}
Hannun, A.~Y.; Rajpurkar, P.; Haghpanahi, M.; Tison, G.~H.; Bourn, C.;
  Turakhia, M.~P.; and Ng, A.~Y. 2019.
\newblock Cardiologist-level arrhythmia detection and classification in
  ambulatory electrocardiograms using a deep neural network.
\newblock \emph{Nature Medicine} 25(1): 65.

\bibitem[{Jetley et~al.(2018)Jetley, Lord, Lee, and Torr}]{jetley2018learn}
Jetley, S.; Lord, N.~A.; Lee, N.; and Torr, P. 2018.
\newblock Learn to Pay Attention.
\newblock In \emph{International Conference on Learning Representations}.

\bibitem[{Karlen et~al.(2013)Karlen, Raman, Ansermino, and
  Dumont}]{karlen2013multiparameter}
Karlen, W.; Raman, S.; Ansermino, J.~M.; and Dumont, G.~A. 2013.
\newblock Multiparameter respiratory rate estimation from the
  photoplethysmogram.
\newblock \emph{IEEE Transactions on Biomedical Engineering} 60(7): 1946--1953.

\bibitem[{McSharry et~al.(2003)McSharry, Clifford, Tarassenko, and
  Smith}]{mcsharry2003dynamical}
McSharry, P.~E.; Clifford, G.~D.; Tarassenko, L.; and Smith, L.~A. 2003.
\newblock A dynamical model for generating synthetic electrocardiogram signals.
\newblock \emph{IEEE Transactions on Biomedical Engineering} 50(3): 289--294.

\bibitem[{Nguyen et~al.(2017)Nguyen, Le, Vu, and Phung}]{nguyen2017dual}
Nguyen, T.; Le, T.; Vu, H.; and Phung, D. 2017.
\newblock Dual discriminator generative adversarial nets.
\newblock In \emph{Advances in Neural Information Processing Systems},
  2670--2680.

\bibitem[{Nilsson et~al.(2005)}]{nilsson2005respiration}
Nilsson, L.; et~al. 2005.
\newblock Respiration can be monitored by photoplethysmography with high
  sensitivity and specificity regardless of anaesthesia and ventilatory mode.
\newblock \emph{Acta Anaesthesiologica Scandinavica} 49(8): 1157--1162.

\bibitem[{Oktay et~al.(2018)Oktay, Schlemper, Folgoc, Lee, Heinrich, Misawa,
  Mori, McDonagh, Hammerla, Kainz et~al.}]{oktay2018attention}
Oktay, O.; Schlemper, J.; Folgoc, L.~L.; Lee, M.; Heinrich, M.; Misawa, K.;
  Mori, K.; McDonagh, S.; Hammerla, N.~Y.; Kainz, B.; et~al. 2018.
\newblock Attention u-net: Learning where to look for the pancreas.
\newblock \emph{arXiv preprint arXiv:1804.03999} .

\bibitem[{Penttil{\"a} et~al.(2001)Penttil{\"a}, Helminen, Jartti, Kuusela,
  Huikuri, Tulppo, Coffeng, and Scheinin}]{penttila2001time}
Penttil{\"a}, J.; Helminen, A.; Jartti, T.; Kuusela, T.; Huikuri, H.~V.;
  Tulppo, M.~P.; Coffeng, R.; and Scheinin, H. 2001.
\newblock Time domain, geometrical and frequency domain analysis of cardiac
  vagal outflow: effects of various respiratory patterns.
\newblock \emph{Clinical Physiology} 21(3): 365--376.

\bibitem[{Pimentel et~al.(2016)Pimentel, Johnson, Charlton, Birrenkott,
  Watkinson, Tarassenko, and Clifton}]{pimentel2016toward}
Pimentel, M.~A.; Johnson, A.~E.; Charlton, P.~H.; Birrenkott, D.; Watkinson,
  P.~J.; Tarassenko, L.; and Clifton, D.~A. 2016.
\newblock Toward a robust estimation of respiratory rate from pulse oximeters.
\newblock \emph{IEEE Transactions on Biomedical Engineering} 64(8): 1914--1923.

\bibitem[{Reiss et~al.(2019)Reiss, Indlekofer, Schmidt, and
  Van~Laerhoven}]{reiss2019deep}
Reiss, A.; Indlekofer, I.; Schmidt, P.; and Van~Laerhoven, K. 2019.
\newblock Deep PPG: large-scale heart rate estimation with convolutional neural
  networks.
\newblock \emph{Sensors} 19(14): 3079.

\bibitem[{Ross et~al.(2019)Ross, Sarkar, Rodenburg, Ruberto, Hungler,
  Szulewski, Howes, and Etemad}]{ross2019toward}
Ross, K.; Sarkar, P.; Rodenburg, D.; Ruberto, A.; Hungler, P.; Szulewski, A.;
  Howes, D.; and Etemad, A. 2019.
\newblock Toward Dynamically Adaptive Simulation: Multimodal Classification of
  User Expertise Using Wearable Devices.
\newblock \emph{Sensors} 19(19): 4270.

\bibitem[{{Sarkar} and {Etemad}(2020{\natexlab{a}})}]{sarkar2020}
{Sarkar}, P.; and {Etemad}, A. 2020{\natexlab{a}}.
\newblock Self-supervised ECG Representation Learning for Emotion Recognition.
\newblock \emph{IEEE Transactions on Affective Computing} 1--1.

\bibitem[{{Sarkar} and {Etemad}(2020{\natexlab{b}})}]{sarkar2019self}
{Sarkar}, P.; and {Etemad}, A. 2020{\natexlab{b}}.
\newblock Self-Supervised Learning for ECG-Based Emotion Recognition.
\newblock In \emph{IEEE International Conference on Acoustics, Speech and
  Signal Processing}, 3217--3221.

\bibitem[{{Sarkar} et~al.(2019){Sarkar}, {Ross}, {Ruberto}, {Rodenbura},
  {Hungler}, and {Etemad}}]{sarkar2019}
{Sarkar}, P.; {Ross}, K.; {Ruberto}, A.~J.; {Rodenbura}, D.; {Hungler}, P.; and
  {Etemad}, A. 2019.
\newblock Classification of Cognitive Load and Expertise for Adaptive
  Simulation using Deep Multitask Learning.
\newblock In \emph{IEEE International Conference on Affective Computing and
  Intelligent Interaction}, 1--7.

\bibitem[{Sayadi, Shamsollahi, and Clifford(2010)}]{sayadi2010synthetic}
Sayadi, O.; Shamsollahi, M.~B.; and Clifford, G.~D. 2010.
\newblock Synthetic ECG generation and Bayesian filtering using a Gaussian
  wave-based dynamical model.
\newblock \emph{Physiological Measurement} 31(10): 1309.

\bibitem[{Sch{\"a}ck et~al.(2017)}]{schack2017computationally}
Sch{\"a}ck, T.; et~al. 2017.
\newblock Computationally efficient heart rate estimation during physical
  exercise using photoplethysmographic signals.
\newblock In \emph{European Signal Processing Conference}, 2478--2481.

\bibitem[{Sch{\"a}fer and Vagedes(2013)}]{schafer2013accurate}
Sch{\"a}fer, A.; and Vagedes, J. 2013.
\newblock How accurate is pulse rate variability as an estimate of heart rate
  variability?: A review on studies comparing photoplethysmographic technology
  with an electrocardiogram.
\newblock \emph{International Journal of Cardiology} 166(1): 15--29.

\bibitem[{Schmidt et~al.(2018)Schmidt, Reiss, Duerichen, Marberger, and
  Van~Laerhoven}]{schmidt2018introducing}
Schmidt, P.; Reiss, A.; Duerichen, R.; Marberger, C.; and Van~Laerhoven, K.
  2018.
\newblock Introducing wesad, a multimodal dataset for wearable stress and
  affect detection.
\newblock In \emph{Proceedings of the International Conference on Multimodal
  Interaction}, 400--408.

\bibitem[{Shelley et~al.(2006)Shelley, Awad, Stout, and
  Silverman}]{shelley2006use}
Shelley, K.~H.; Awad, A.~A.; Stout, R.~G.; and Silverman, D.~G. 2006.
\newblock The use of joint time frequency analysis to quantify the effect of
  ventilation on the pulse oximeter waveform.
\newblock \emph{Journal of Clinical Monitoring and Computing} 20(2): 81--87.

\bibitem[{WHO(2017)}]{who_cvds}
WHO. 2017.
\newblock Cardiovascular Diseases.
\newblock
  \url{https://www.who.int/news-room/fact-sheets/detail/cardiovascular-diseases-(cvds)}.
\newblock (Accessed on 07/10/2020).

\bibitem[{{Zhang}, {Zhou}, and {Zeng}(2017)}]{heartid}
{Zhang}, Q.; {Zhou}, D.; and {Zeng}, X. 2017.
\newblock HeartID: A Multiresolution Convolutional Neural Network for ECG-Based
  Biometric Human Identification in Smart Health Applications.
\newblock \emph{IEEE Access} 5: 11805--11816.

\bibitem[{Zhu et~al.(2019{\natexlab{a}})Zhu, Ye, Fu, Liu, and
  Shen}]{zhu2019electrocardiogram}
Zhu, F.; Ye, F.; Fu, Y.; Liu, Q.; and Shen, B. 2019{\natexlab{a}}.
\newblock Electrocardiogram generation with a bidirectional LSTM-CNN generative
  adversarial network.
\newblock \emph{Scientific Reports} 9(1): 1--11.

\bibitem[{Zhu et~al.(2017)Zhu, Park, Isola, and Efros}]{CycleGAN2017}
Zhu, J.-Y.; Park, T.; Isola, P.; and Efros, A.~A. 2017.
\newblock Unpaired image-to-image translation using cycle-consistent
  adversarial networks.
\newblock In \emph{Proceedings of the IEEE International Conference on Computer
  Vision}, 2223--2232.

\bibitem[{Zhu et~al.(2019{\natexlab{b}})Zhu, Tian, Wong, and
  Wu}]{zhu2019learning}
Zhu, Q.; Tian, X.; Wong, C.-W.; and Wu, M. 2019{\natexlab{b}}.
\newblock Learning Your Heart Actions From Pulse: ECG Waveform Reconstruction
  From PPG.
\newblock \emph{bioRxiv} 815258.

\end{thebibliography}
\end{document}